\documentclass{article}

\usepackage{microtype}
\usepackage{graphicx}
\usepackage{subfigure}
\usepackage{booktabs} 
\usepackage{float}

\usepackage{braket}
\usepackage{amsfonts}
\usepackage[shortlabels]{enumitem}
\usepackage{amsmath}
\usepackage[makeroom]{cancel}
\usepackage{dsfont}
\usepackage{multirow}

\DeclareMathOperator{\tr}{tr}

\usepackage{hyperref}

\usepackage[accepted]{icml2018}

\icmltitlerunning{Learning Quantum Graphical Models Using Constrained Gradient Descent on the Stiefel Manifold}

\begin{document}

\twocolumn[
\icmltitle{Learning Quantum Graphical Models using Constrained Gradient Descent on the Stiefel Manifold}

\icmlsetsymbol{equal}{*}

\begin{icmlauthorlist}
\icmlauthor{Sandesh Adhikary $^*$}{to}
\icmlauthor{Siddarth  Srinivasan $^*$}{to}
\icmlauthor{Byron Boots}{to}
\end{icmlauthorlist}

\icmlaffiliation{to}{College of Computing, Georgia Institute of Technology, Atlanta, GA, USA}

\icmlcorrespondingauthor{Sandesh Adhikary}{sadhikary6@gatech.edu}

\icmlkeywords{Quantum, Kraus operator, Stiefel Manifold, Machine Learning, ICML}

\vskip 0.3in
]

 \printAffiliationsAndNotice{\icmlEqualContribution} 

\begin{abstract}
Quantum graphical models (QGMs) extend the classical framework for reasoning about uncertainty by incorporating the quantum mechanical view of probability. Prior work on QGMs has focused on hidden quantum Markov models (HQMMs), which can be formulated using quantum analogues of the sum rule and Bayes rule used in classical graphical models. Despite the focus on developing the QGM framework, there has been little progress in learning these models from data. The existing state-of-the-art approach randomly initializes parameters and iteratively finds unitary transformations that increase the likelihood of the data~\cite{srinivasan2018}. While this algorithm demonstrated theoretical strengths of HQMMs over HMMs, it is slow and can only handle a small number of hidden states. In this paper, we tackle the learning problem by solving a constrained optimization problem on the Stiefel manifold using a well-known retraction-based algorithm. We demonstrate that this approach is not only faster and yields better solutions on several datasets, but also scales to larger models that were prohibitively slow to train via the earlier method.
\end{abstract}

\section{Introduction and Related Work}
Classical probabilistic graphical models provide a principled framework for Bayesian reasoning, and there has been much interest in extending this framework to develop quantum graphical models (QGMs) by incorporating the quantum mechanical view of probability \cite{leifer2008quantum, yeang2010probabilistic, leifer2013towards, warmuth2014bayesian}. There has also been a focus on using hidden quantum Markov models (HQMMs) to model stochastic processes \cite{monras2010hidden, clark2015hidden}, and recent work \cite{srinivasan2018} brought HQMMs into the QGM framework by showing how they can be formulated using quantum analogues of the sum and Bayes rule. A major motivation for investigating QGMs is the promise of a more general and expressive class of probabilistic models. \citet{srinivasan2018} recently showed that HMMs are a subset of HQMMs and used several synthetic datasets to empirically demonstrate the theoretical advantages of HQMMs over HMMs. Note that these proposals are agnostic about the development of a quantum computer; their focus is to simply use the mathematical formalism of quantum mechanics (QM) for probabilistic reasoning.

 Yet despite the focus on developing the QGM framework, there has been little progress in learning these models from data. QGMs are typically parameterized by a set of Kraus operators $\{\hat{K}_i\}$ satisfying the constraint $\sum_i \hat{K}_i^\dagger \hat{K}_i = \mathds{I}$. When stacked vertically to form a matrix $\kappa$, we have $\kappa^\dagger \kappa = \mathds{I}$. The existing approach \citep{srinivasan2018} yields feasible parameters by starting with an initial guess $\kappa^*$ and iteratively finding unitary transformations that increase the likelihood of the data. However, this method is inefficient, often gets trapped in poor optima, and can only handle a small number of hidden states. Our primary contribution in this paper is the application of an alternative approach to the learning problem: since $\kappa$ lies on the Stiefel manifold \cite{stiefel-original,Edelman1998}, we can directly learn feasible parameters by constraining gradient updates to lie on the manifold using a well-known retraction-based algorithm \cite{Wen2013}. We show that this approach is faster, finds better optima, and can handle more hidden states compared to the previous method.

This paper is structured as follows: in section 2 we provide an overview of the QGM framework as developed by \citet{srinivasan2018, NIPS2018_8235} and show that the model parameters are matrices on the Stiefel manifold. In section 3, we describe the retraction-based optimization algorithm  to learn model parameters, and finally present experimental results for learning HQMMs on several datasets in section 4.

\vspace{-3mm}
\section{Quantum Graphical Models}
\subsection{State Representation}
In a classical probabilistic graphical model (PGM), the probability mass function of a discrete random variable $X$ can be represented by a vector $\vec{x} \in \mathds{R}^n$, called a \emph{belief state}. The entries must satisfy  $\|\vec{x}\|_1 = 1$, and each entry corresponds to the probability of the corresponding system state. The analogous representation of belief in a quantum graphical model is the Hermitian, positive semi-definite $\emph{density matrix}$ $\hat{\rho}$ which much satisfy $\text{tr}(\hat{\rho}) = 1$. Here, $\hat{\rho}  \in \mathds{C}^{n\times n}$ and each of its diagonal entries represents the probability of the corresponding system state. In quantum information theory, the off-diagonal elements of the density matrix represent quantum coherences and entanglement, and do not have a straightforward interpretation. For our purposes, we can think of the off-diagonal entries as encoding additional information about the probability distribution.

\paragraph{Joint Distributions and Marginalization} If we represent the distributions of random variables $A$ and $B$ with the density matrices $\hat{\rho}_A$ and $\hat{\rho}_B$, the joint state can be represented using the tensor product $\hat{\rho}_A \otimes \hat{\rho}_B = \hat{\rho}_{AB}$. Similarly, given a joint distribution (which may not factor as a product of states), we can compute the marginal distribution using the \emph{partial trace} operation $\hat{\rho}_A = \text{tr}_B(\hat{\rho}_{AB})$ (see \citet{NIPS2018_8235} and \citet{nielsen_chuang_2010} for details).

\begin{figure}[h]
\begin{center}
\includegraphics[scale=0.5]{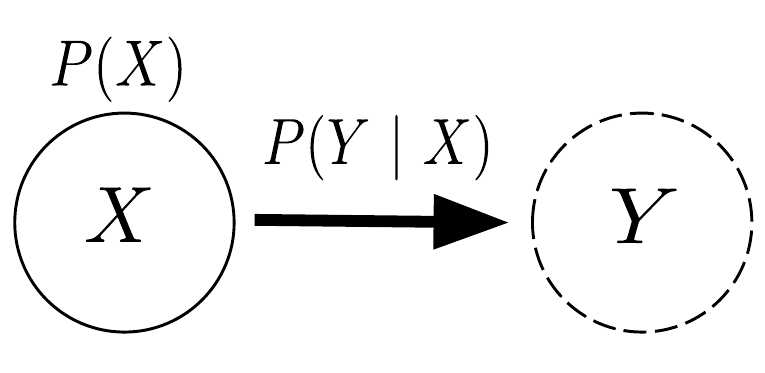}
\end{center}
\caption{A simple graphical model of two random variables X, Y}\label{simplegm}
\end{figure}
\subsection{Inference Using QGMs}\label{sectwo}

Consider the simple graphical model (shown in Figure \ref{simplegm}) of two random variables $X$ and $Y$ with prior $P(X) = \vec{x}$ and likelihood  $P(Y|X)$ written as a column-stochastic matrix ${\bf A}$ (non-negative matrix whose columns sum to 1). We can use the regular sum rule to compute $P(Y) = \sum_x P(Y|x)P(x) = {\bf A}\vec{x}$, and use Bayes rule  to compute $P(X|y) = \frac{P(y|X)P(X)}{P(y)} = \frac{\text{diag}({\bf A}_{(y,:)})\vec{x}}{\mathds{1}^T\text{diag}({\bf A}_{(y,:)})\vec{x}}$. Here, $\text{diag}({\bf A}_{(y,:)})$ is a diagonal matrix constructed using the row of ${\bf A}$ corresponding to  observation $y$, and the denominator renormalizes the numerator to  yield the posterior.

In general, column-stochastic matrices constitute the class of matrices that can be used to update a belief state in a classical PGM. When updating the belief state of some evolving random variable, we can think of the column-stochastic matrix as a \emph{transition matrix} encoding dynamics. By the sum rule, we have $\vec{x}_{t+1} = {\bf A}\vec{x}_t$, and such a matrix is guaranteed to take one valid belief state to another.

\paragraph{Quantum Channels} Naturally, the next question is: what are the valid operations that can evolve a \emph{quantum} belief state represented as a density matrix $\hat{\rho}$ to another valid density matrix? This can be achieved through completely-positive, linear maps known as quantum channels $\mathcal{K}$, which must fulfill the following three conditions:
\begin{enumerate}[I.]
    \vspace{-2mm}
    \item Hermiticity Preservation (HP):
    $\mathcal{K}(\hat{\rho}) = \mathcal{K}(\hat{\rho})^\dagger$
    \vspace{-2mm}
    \item Complete Positivity (CP):
    $
    [\mathcal{K} \otimes \mathbb{I}_m] (\hat{\rho} \otimes \hat{\rho}_m) \geq 0
    $.
    \vspace{-2mm}
    \item Trace Preservation (TP):
    $
    \tr \left[ \mathcal{K}(\hat{\rho}) \right] = 1
    $
    \vspace{-2mm}
\end{enumerate}

The HP and TP conditions follow from the fact that all valid density matrices are Hermitian and have unit trace. Since density matrices are also positive semi-definite, $\mathcal{K}$ must preserve this property. However, \textit{complete positivity} (CP) is a stricter condition requiring that any joint operation $(\mathcal{K} \otimes \mathbb{I}_m)$ (i.e., where $\mathcal{K}$ operates on  the primary system and the identity map $\mathds{I}_m$ is applied on some arbitrary auxillary density matrix $\hat{\rho}_m$ of dimension $m \times m$) also produce a positive semi-definite matrix \cite{Gheondea2010}.

\paragraph{Parameterizing Quantum Channels} So what are the class of matrices we can use to implement a quantum channel? Quantum channels and the more general CP maps have been studied extensively in quantum information \cite{nielsen_chuang_2010,Gheondea2010,Verstraete2004}, and can be parameterized using using a set of matrices $\{\hat{K}_w\}$ called Kraus operators \cite{Kraus1971} via the following operator-sum representation  (the $\dagger$ represents the complex-conjugate transpose):
\begin{align}
    \mathcal{K}(\hat{\rho}) = \sum_w \hat{K}_w \hat{\rho} \hat{K}_w^{\dagger}
    \label{eq:hqmm_kraus-rep}
\end{align}
Any operation $\mathcal{K}$ that can be expressed in the form of Eq.~\ref{eq:hqmm_kraus-rep} is guaranteed to be both HP and CP \cite{Gheondea2010, Pillis1967}. Since density matrices are Hermitian PSD, applying Kraus operators as shown above will naturally preserve the HP and CP constraints. The TP requirement on $\mathcal{K}$ however must be explicitly preserved, and can be expressed as the following constraint on Kraus operators:
\begin{equation}
    \sum_w {\hat{K}_w}^\dagger \hat{K}_w = \mathbb{I}
    \label{eq:kraus-rep-tp}
\end{equation}
Note that if the set $\{\hat{K}_w\}$ only contains a single square Kraus operator, the channel $\mathcal{K}$ reduces to the simplest valid operation on a density matrix: unitary evolution where $\mathcal{K}(\hat{\rho}) =  \hat{U}\hat{\rho}\hat{U}^\dagger$ with $\hat{U}^\dagger \hat{U} = \mathds{I}$. Kraus operators are a convenient representation of quantum channels as the  constraints I-III can easily be satisfied by construction.
  
  Now, observe that if we vertically stack all the Kraus operators in our  set to form the matrix $\kappa = \begin{bmatrix}\hat{K}_1 & \hat{K}_2  &  \ldots & \hat{K}_N \end{bmatrix}^T$, it will satisfy $\kappa^\dagger \kappa = \mathds{I}$ \cite{Oza2009}. If each Kraus operator is of dimension $n$, then we have $\kappa \in \mathds{C}^{nN \times n}$ and the set of such `tall and skinny' matrices with $n$ orthonormal columns in $\mathds{C}^{nN}$ constitute the Stiefel manifold. Conversely, if we start with a $nN \times n$ matrix $\kappa$ on the Stiefel manifold of $\mathds{C}^{nN}$, we can partition it into $N$ blocks of $n \times n$ matrices and these blocks would constitute a set of valid Kraus operators. This is precisely the result that will allow us to directly optimize a $\kappa$ matrix on the Stiefel manifold \cite{Absil2007,Edelman1998} and recover the Kraus operators from it.

Having described the valid operations on a quantum belief state $\hat{\rho}$, we consider the QGM analogue of the graphical model shown in Figure  \ref{simplegm} and present the basic operations: the quantum sum rule and Bayes rule.

\paragraph{Sum Rule}  Given a prior encoded in a density matrix $\hat{\rho}_X$ and a set of Kraus operators $\{\hat{K}_w\}$ that encode how $X$ probabilistically affects $Y$, the quantum sum rule is as follows:
\begin{equation}
    \hat{\rho}_Y = \sum_w \hat{K}_w \hat{\rho}_X \hat{K}_w^\dagger
\end{equation}
In classical PGMs that model some kind of dynamics, the sum rule is the operation used to model `transitions' or evolution of the belief state; in the quantum case, these dynamics are encoded in the Kraus operators. 

\paragraph{Bayes Rule} The quantum analogue of Bayes rule for conditioning on observation $y$ is given as follows:
\begin{equation}
    \hat{\rho}_{X|y} = \frac{\hat{K}_y \hat{\rho}_X \hat{K}_y^\dagger}{\text{tr}\left(\sum_y \hat{K}_y \hat{\rho}_X \hat{K}_y^\dagger\right)}
\end{equation}
The numerator produces a matrix whose diagonal entries encode the joint probability of the observation $y$ and the corresponding system state, and the denominator simply takes the trace of the numerator to `sum out' the probabilities of the system states. Just like the classical Bayes rule, this denominator term $\text{tr}\left(\sum_y \hat{K}_y \hat{\rho}_X \hat{K}_y^\dagger\right)$ gives the probability of the observation $y$ and renormalizes the density matrix.

\subsection{Hidden Quantum Markov Models}
\begin{figure}[h]
    \centering
    \includegraphics[scale = 0.6]{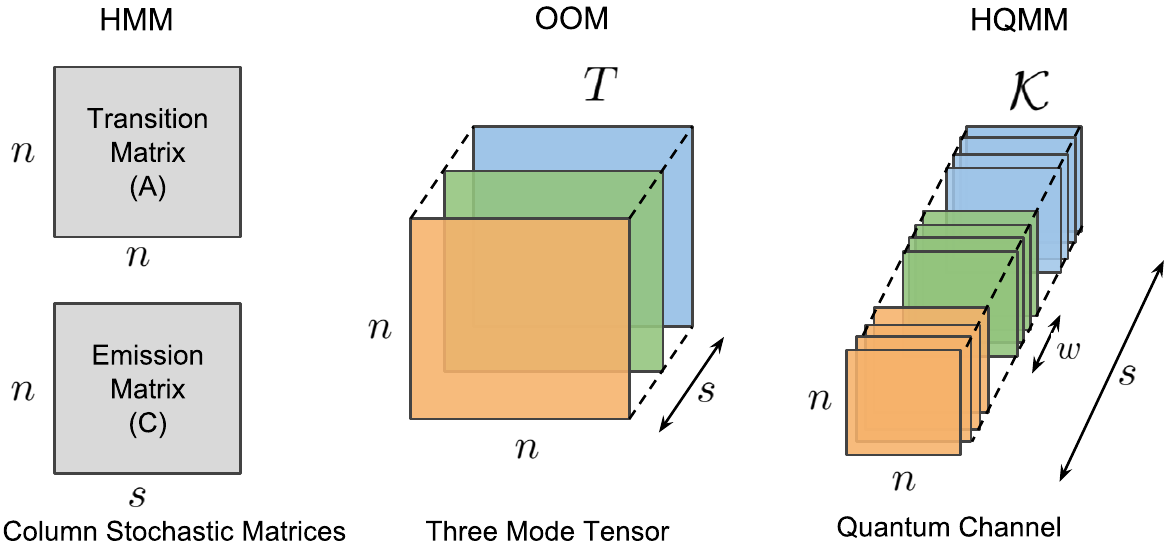}
    \caption{\textbf{Parameterizations of HMMs, OOMs, and HQMMs.} HMMs have a transition and emission matrix, OOMs consist of a tensor $T$ with each of its $s$ fibers corresponding to a unique output. HQMMs have a similar tensor structure, but consist of $s$ sub-tensors with $w$ fibers for each unique observation.}
    \label{fig:tensor_fig}
\end{figure}

Since we will be working with HQMMs as our particular choice of QGMs, we first give a brief overview of hidden Markov models (HMMs). An HMM can be composed by the repeated application of the sum rule for transition dynamics and Bayes rule for conditioning on observation, using column-stochastic matrices ${\bf A}$ and ${\bf C}$ respectively:
\begin{equation}
\begin{split}
    \vec{x}_{t}' &= {\bf A}\vec{x}_{t-1} \\
    \vec{x}_{t} &= \frac{\text{diag}({\bf C}_{(y,:)})\vec{x}_{t}'}{\mathds{1}^T \text{diag}({\bf C}_{(y,:)})\vec{x}_{t}'}
\end{split}\label{hmm:up}
\end{equation}
\begin{figure}[t]
\label{fig:hmm}
  \centering
    \includegraphics[width=0.3\textwidth]{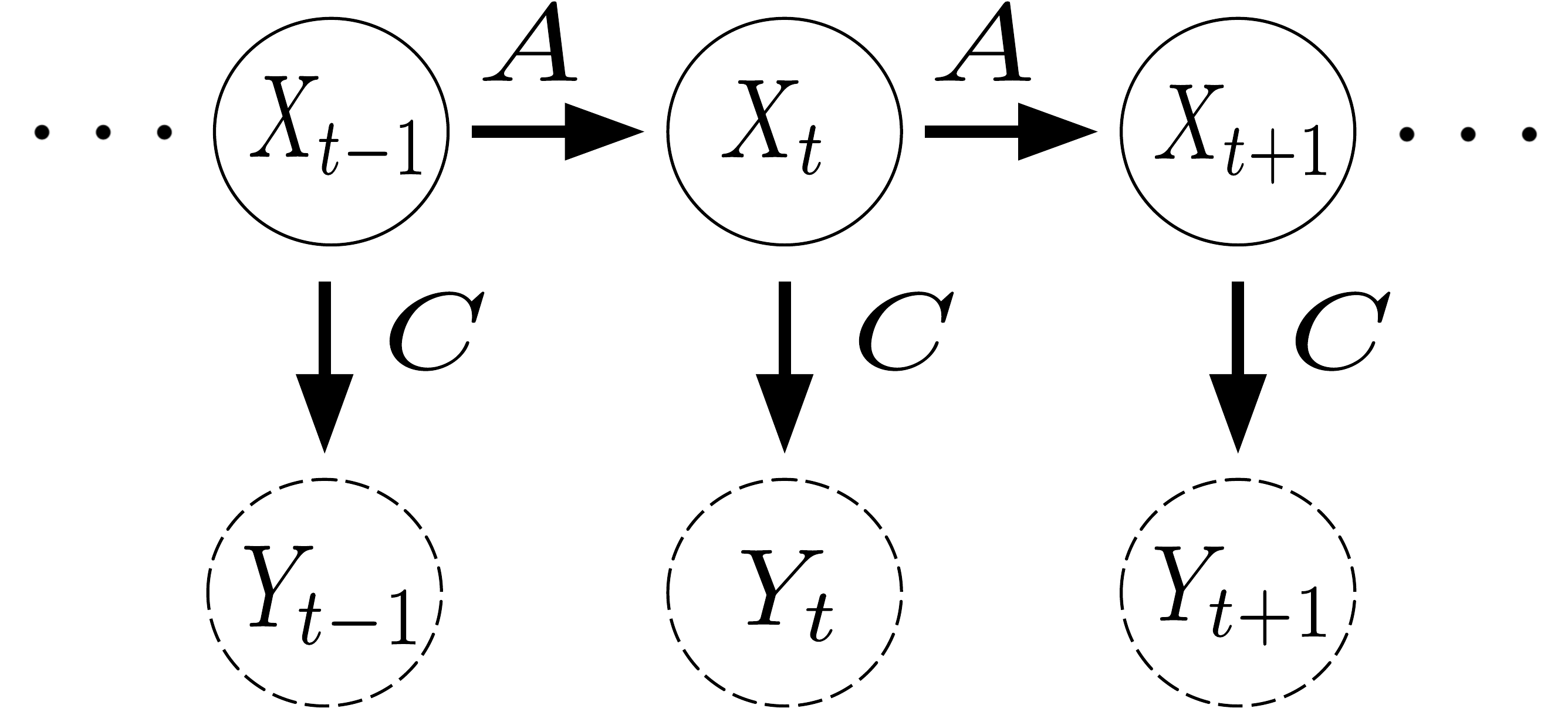}
      \caption{A Hidden Markov Model}
\end{figure} 
However, this operation can be condensed into a single update (parameterized by a 3-mode tensor as shown in Figure \ref{fig:tensor_fig}) for a given observation $y$ using the observable operator model (OOM) representation \cite{jaeger2000observable}:
\begin{equation}
    \vec{x}_{t} = \frac{\text{diag}({\bf C}_{(y,:)}){\bf A}\vec{x}_{t-1}}{\mathds{1}^T \text{diag}({\bf C}_{(y,:)}){\bf A}\vec{x}_{t-1}} = \frac{{\bf T}_y\vec{x}_{t-1}}{\mathds{1}^T {\bf T}_y\vec{x}_{t-1}}\label{oom}.
\end{equation}

Analogously, an HQMM can be composed by  the repeated application of the quantum sum rule and quantum Bayes rule encoded using the sets of Kraus operators $\{ \hat{K}_w\}$ and $\{ \hat{K}_y \}$  respectively:
\begin{equation}
\begin{split}
    \hat{\rho}_{t}' &= \sum_w \hat{K}_w \hat{\rho}_{t-1} \hat{K}_w^\dagger \\ \vspace{-2mm}
    \hat{\rho}_t &= \frac{\hat{K}_y \hat{\rho}_t' \hat{K}_y^\dagger}{\text{tr}\left(\sum_y \hat{K}_y \hat{\rho}_X \hat{K}_y^\dagger\right)}
\end{split}\label{hqmm:sep}
\end{equation}\vspace{-6mm}

As in the classical case, we can condense the previous updates into a single update for a given observation $y$ by  setting $\hat{K}_{y,w} = \hat{K}_y\hat{K}_w$:
\begin{equation}
    \hat{\rho}_t = \frac{\sum_{w}\hat{K}_{y,w} \hat{\rho}_{t-1} \hat{K}_{y,w}^\dagger}{\text{tr}\left(\sum_{w} \hat{K}_{y,w} \hat{\rho}_{t-1} \hat{K}_{y,w}^\dagger\right)} \label{hqmm:joined}
\end{equation}
As mentioned earlier, the numerator term produces an unnormalized density matrix whose trace gives the probability of the observation $y$, a fact we will use to construct a likelihood loss for a sequence of observations. The HQMM is characterized by the dimension of the latent state (i.e., the density matrix) $n$, the number of outputs $s$, and the number of terms in the sum in the numerator of Equation \ref{hqmm:joined}, $w$ (corresponds to the dimension of an `environment' variable). Thus, there are $sw$ Kraus operators of dimension $n \times n$, giving a total of $n^2sw$ parameters that form the quantum channel $\mathcal{K}$. As shown in Figure~\ref{fig:tensor_fig}, a quantum channel can also be viewed as a tensor similar to OOMs. However, quantum channels encode the dynamics associated with every observable in a separate $n \times n \times w$ tensor, as opposed to a single matrix in OOMs. This structure provides HQMMs with an additional hyperparameter $w$ not available in HMMs and OOMs.

The HQMM update rule in Equation \ref{hqmm:joined} is a generalization of the HMM update rule in Equation \ref{oom}; \citet{srinivasan2018} showed how the matrices $\{{\bf T}_y\}$ that parameterize HMMs can be encoded in Kraus operators $\{\hat{K}_{y,w}\}$ so that both classical and quantum belief states are updated identically. Thus, HQMMs are a more expressive class of models that contain HMMs as a subset.

\section{The Learning Problem}
\subsection{The Loss Function}
Since QGMs are probabilistic models, the negative log-likelihood of the data is a natural loss function to use. Just as in PGMs, this will involve computing a joint distribution using the structure of the graphical model and marginalizing over the latent variables to obtain the likelihood of the observed variables. Here, there is an important distinction between PGMs and QGMs. In a PGM, when we compute the joint distribution of a single latent variable and some observed variables, we might use an update like the numerator of the second update in Equation \ref{hmm:up}: let $\vec{z} = \text{diag}({\bf C}_{(y_n,:)})\cdots \text{diag}({\bf C}_{(y_1,:)})\vec{x}$ so $\vec{z}$ contains the joint probability $P(X,y_1,\ldots,y_n)$. Since the observations are encoded in diagonal matrices $\text{diag}({\bf C}_{(y_i,:)})$ that commute, the order in which the joint distribution is constructed does not matter. In a QGM, constructing such a joint involves multiplying Kraus operators as $\hat{\rho}_z=\hat{K}_{y_n}\cdots\hat{K}_{y_1}\hat{\rho}_X\hat{K}_{y_1}^\dagger\cdots\hat{K}_{y_n}^\dagger$. In the special case where we obtain these Kraus operators by converting a column-stochastic matrix according to the scheme proposed by \citet{srinivasan2018}, the Kraus operators are diagonal and commute, and we get the same result as the classical case. However, they  \emph{need not commute in general}, so the order in which we construct the joint distribution matters! 

This is only relevant for QGMs where we need to construct the joint distribution of a latent variable connected to multiple observed variables. For instance, if we considered the QGM version of a Naive Bayes model, in general, the order in which the features are used to compose the joint will matter, breaking the central assumption about independence of features. A full treatment of the consequences of the non-commutativity of Kraus operators in a quantum Naive Bayes model (and QGMs in general) is beyond the scope of this paper, although it is worthy of further study. 

Our experiments focus on HQMMs, and the latent variable in an HQMM at any given time-step is only tied to one observed variable at that time-step so we can simply construct the joint distribution of the sequence of observations in the order that they are observed. This yields the following loss function for an HQMM, parameterized by the set of Kraus operators  $\{\hat{K}_{y,w}\}$ \citep{srinivasan2018}:
\vspace{-1mm}
\begin{equation}
    \mathcal{L} = -\ln \text{tr}\left(\sum_{w}\hat{K}_{y_n,w}\ldots \left(\sum_{w}\hat{K}_{{y_1},w}\hat{\rho}_0\hat{K}_{{y_1},w}^\dagger\right)\ldots \hat{K}_{y_n,w}^\dagger\right) 
    \label{eq:loss}
\end{equation}
Note that using a negative log-likelihood loss to learn model parameters requires that the parameters be properly constrained to produce valid QGMs so that increasing the log-likelihood of some observations requires moving probability mass away from other possible observations; otherwise we could  simply arbitrarily scale the parameters to maximize the log-likelihood. Having parameterized the loss in terms of Kraus operators, we now proceed to the task of minimizing the loss while ensuring that the learned Kraus operators satisfy the TP constraint $\sum_{y,w} \hat{K}^\dagger_{y,w} \hat{K}_{y,w} = \mathds{I}$.

\subsection{The Optimization Algorithm}
As discussed in Section \ref{sectwo}, the problem of learning a set of $N$ trace-preserving $n\times n$ Kraus operators can equivalently be framed as one of learning a matrix $\kappa \in \mathds{C}^{nN\times n}$ with  orthonormal columns on the Stiefel manifold, where $\kappa$ can be partitioned row-wise into the $N$ Kraus operators that parameterize the QGM. Both the previous approach and the approach in this paper begin with an initial guess $\kappa_0$ with a pre-determined partitioning into the Kraus operators we wish to learn, and iteratively make changes to the guess to maximize the log-likelihood (which is a function of those partitioned blocks of Kraus operators).

\paragraph{The Previous Approach} Since $\kappa$ is a matrix with orthonormal columns, any initial guess $\kappa_0$ is a unitary transformation away from the true $\kappa^*$ that maximizes the log-likelihood. The existing method \cite{srinivasan2018} tries to find this unitary transformation, but it does not do so directly. It uses the fact that any unitary matrix can be decomposed into a series of Givens rotations to instead iteratively find such rotations that locally increase the log-likelihood. Such a complex Givens rotation only has 4 parameters, so a local rotation is easy enough to find; however, a Givens rotation only changes two rows of $\kappa$ at a time, making this approach too slow for learning large $\kappa$ matrices.

\paragraph{A Retraction-Based Optimization Algorithm} Instead of the previous indirect approach to learning model parameters, we propose directly learning the parameters in matrix $\kappa$ using a gradient-based algorithm. Note that since $\mathcal{L}$ is a real-valued function of complex Kraus operators, the direction of steepest descent corresponds to the gradient with respect to the complex conjugate of the Kraus operators \cite{Hjorungnes2007}. However, $\kappa$ must satisfy $\kappa^\dagger \kappa = \mathds{I}$, so we cannot simply take the gradient of the loss $\mathcal{L}$ with respect to the parameters and step along it -- this can easily move us off the Stiefel manifold. The class of retraction-based algorithms provides a means of moving along the gradient direction $G$, while staying on the manifold of feasible parameters.

\begin{figure}[h]
\label{fig:retraction}
  \centering
    \includegraphics[scale = 0.65]
    {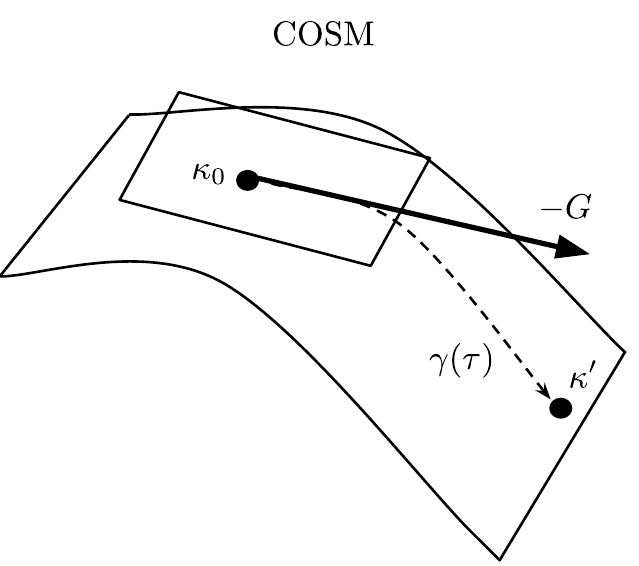}
    \caption{\textbf{An Illustration of Retraction-based Optimization on the Stiefel Manifold}: Starting from initial parameters $\kappa_0$, we compute a retraction $\gamma(\tau)$ of the negative gradient of the loss function $-G$ onto the Stiefel manifold. Following the trajectory $\gamma(\tau)$ guarantees staying on the manifold. 
    }\label{retraction}
\end{figure}

Given a gradient $G$ of the loss function $\mathcal{L}$ with respect to parameters $\kappa$, we would ideally like to move along a trajectory $\gamma(\tau)$ for some step size $\tau$, such that moving along this curve corresponds to stepping along the direction of the gradient while staying on the manifold.
We can achieve this through \textit{retractions}, which smoothly map a point on the tangent bundle of the manifold onto the manifold itself, while preserving the direction of the gradient at that point \cite{Absil2007}, as illustrated in Figure \ref{retraction}. The gradient $G$ of $\mathcal{L}$ corresponds to a vector on the tangent bundle of this manifold, so intuitively we can imagine a retraction as wrapping the direction of $G$ onto the surface of the manifold, providing a feasible path for curvilinear descent. 

While there are a number of algorithms for constrained optimization on the Stiefel manifold, we use the state-of-the-art algorithm proposed by \citet{Wen2013} for its ease of implementation. They present the following Crank-Nicolson-like update $\gamma(\tau)$ on the Stiefel manifold with respect to an initial feasible solution $\kappa_0$:
\begin{equation}
\gamma(\tau) = \left(\mathbb{I} + \frac{\tau}{2} A\right)^{-1} \left(\mathbb{I} - \frac{\tau}{2} A\right) \kappa_0,
\label{eq:stiefel_curve}
\end{equation}
where $A$ is a skew-symmetric matrix defined as:
\begin{align}
A = G \kappa_0^\dagger - \kappa_0 G^\dagger
\label{eq:stiefel_skewsym}
\end{align}

If the path $\gamma(\tau)$ is the direction of steepest descent to feasibly optimize Equation ~\ref{eq:loss}, it must fulfill two criteria. First, when we take no steps along the path (i.e., when $\tau = 0$), we should stay at the initial point $\kappa_0$ on the manifold, and the curve $\gamma(\tau)$ should point in the direction of the gradient with respect to $\mathcal{L}$. This can be easily verified as $\gamma(0) = \kappa_0$ and $\gamma'(0) = - G$ \cite{Wen2013}. Second, moving along $\gamma(\tau)$ should not move us off the manifold regardless of the step-size $\tau$. Since the path $\gamma(\tau)$ in  Equation \ref{eq:stiefel_curve} is the Cayley transform of the skew-symmetric matrix $A$ applied to $\kappa_0$, we know $\gamma(\tau)^\dagger \gamma(\tau) = \kappa_0^\dagger \kappa_0 = \mathds{I}$. This ensures that as long as we begin with a point on the Stiefel manifold, the update in Equation \ref{eq:stiefel_curve} will keep the point on the manifold for any $\tau$ and $G$.

In practice, the matrix $\kappa \in \mathds{C}^{nN \times n}$ is `tall and skinny' and we would like to avoid computing the inverse $\left(\mathbb{I} + \frac{\tau}{2} A\right)^{-1}$ where $A \in \mathds{C}^{nN\times nN}$. Hence, we use the following equivalent but computationally more efficient expression for $\gamma(\tau)$ recommended by \citet{Wen2013}:
\begin{equation}
\gamma(\tau) = \kappa - \tau U \left( \mathbb{I} + \frac{\tau}{2} V^\dagger U \right)^{-1} V^\dagger \kappa,
\label{eq:stiefel_smw_curve}
\end{equation}
where $U = [G~|~\kappa_0]$ and $V = [\kappa_0~|-G]$. This only  requires inverting a $2n\times 2n$ matrix.

This method can be combined with a gradient descent scheme (summarized in Algorithm \ref{alg:learn_hqmm}) to learn feasible parameters for any QGM parameterized by $N$ Kraus operators with $\kappa \in \mathds{C}^{nN \times n}$.

\begin{algorithm}[h]
  \caption{Learning QGMs using Constrained Optimization on the Stiefel Manifold}
  \label{alg:learn_hqmm}
  \textbf{Input:} Training data $Y \in \mathds{N}^{M \times \ell}$, where $M$ is the $\#$ of data points and $\ell$ is the $\#$ of observed variables in  the QGM\\
  \textbf{Hyperparameters:} $\mathbf{\tau}$ (learning rate), $\alpha$: (learning rate decay), $B$ (number of batches), $E$ (number of epochs)\\
  \textbf{Output:} Kraus operators $\{\hat{K}_i\}_{i=1}^N$
\begin{algorithmic}[1]
\STATE \textbf{Initialize:} Complex orthonormal matrix on Stiefel manifold $\kappa \in \mathds{C}^{nN \times n}$ and partition into Kraus operators  $\{\hat{K}_i\}_{i=1}^N$, with  $\hat{K}_i \in \mathds{C}^{n\times n}$
    
  \FOR{$epoch$ = 1: $E$ }
  \STATE Partition training data $Y$ into $B$ batches $\{Y_b\}$
    \FOR{$b$ = 1:$B$}
       \STATE Compute gradient $G_i \gets$ $\frac{\partial \mathcal{L}}{\partial \hat{K}_i^*}$ for batch $Y_b$ and loss function $\mathcal{L}$ for the QGM
       \STATE Compute $\frac{\partial \mathcal{L}}{\partial \kappa} = G \leftarrow \begin{bmatrix}G_1 &\cdots & G_N \end{bmatrix}^T$
       \STATE Construct $U \leftarrow [~G~|~\kappa~]$, $ V \leftarrow [~\kappa~|~-G~]$
       \STATE Update $\kappa \leftarrow \kappa - \tau U \left( \mathbb{I} + \frac{\tau}{2} V^\dagger U \right)^{-1} V^\dagger \kappa$
    \ENDFOR
    \STATE Update learning rate $\tau = \alpha \tau$
    \STATE Re-partition $\kappa$ into $\{\hat{K}_i\}$
  \ENDFOR
  
  \STATE \textbf{return} $\{\hat{K}_i\}$
\end{algorithmic}
\end{algorithm}

\vspace{-4mm}
\section{Experimental Results}
To show the superior performance of the retraction-based algorithm for  constrained optimization on the Stiefel manifold ({\bf COSM}) over the previous Givens Search ({\bf GS}) method, we evaluate their accuracy and run-time on three datasets. We use HQMMs as our choice of QGMs for ease of comparison with the previous method. We first compare the two learning algorithms on two synthetic datasets used by \citet{srinivasan2018} (code obtained from Github) -- the first generated by an HQMM and the second by an HMM. We also show the scalability of the COSM approach on a real-world dataset, on which the GS approach is prohibitively slow to train.

\paragraph{Training} For all our HQMMs, we use the log-likelihood loss function from Equation \ref{eq:loss}. We initialize the latent state $\hat{\rho}_0$ as a random Hermitian positive semi-definite matrix. To compute the gradient of the loss function with respect to the complex conjugate of the Kraus operators, we use the Autograd package\footnote{\texttt{https://github.com/HIPS/autograd}} since it can handle derivatives with respect to complex-valued parameters. The gradients of the Kraus operators are then vertically stacked to construct the gradient $\tilde{G}$ of the matrix $\kappa$. Before using the gradient to construct the update to $\kappa$ (as in Line 7 of Algorithm  \ref{alg:learn_hqmm}), we renormalize it by its $2-$norm as $G = \tilde{G}/\|\tilde{G}\|_2$, and use momentum \cite{Rumelhart1998,Qian1999} to update $G = (\beta)\cdot G_{old} + (1-\beta)G$ with $\beta = 0.9$. We renormalize the gradient with momentum by its 2-norm again, so the magnitude of the update is entirely controlled by step-size. We refer to HQMMs using the tuple $(n,s,w)-$HQMM,  where $n$ is the number of hidden states,  $s$ is the number of possible observations,  and $w$ is the dimension of an `environment' variable. Consequently, for an $(n,s,w)-$HQMM we have $N=sw$ and $\kappa \in \mathds{C}^{nsw \times n}$. We also provide the performance of HMMs trained using the Baum-Welch Expectation-Maximization ({\bf EM}) algorithm for reference. 

All experiments were performed a desktop computer with 8 Intel Core i7-7700K 4.20 GHz CPUs, and 31.3 GB RAM. All models are trained in MATLAB, but the gradient computation happens in Python. 

\paragraph{Metrics} On the synthetic HQMM and HMM datasets, we use a scaled log-likelihood \cite{noomreport, srinivasan2018} independent of sequence length called description accuracy: $DA = f\left(1 + \frac{\log_s P(Y|\mathds{D})}{\ell}\right)$, where $f(\cdot)$ squishes the log-likelihood from $(-\infty, 1]$ to $(-1,1)$. When $DA=1$, the model predicted the sequence with perfect accuracy, and when  $DA>0$, the model performed better than random. The error bars represent one standard deviation of the DA scores across many test samples. On the real-world dataset, we report the average accuracy for a classification problem.

\subsection{Synthetic HQMM Data}
For our first experiment, we generated data using a synthetic HQMM with 2 hidden states and 6 possible outputs. This model is fundamentally quantum mechanical; the  data generation process is inspired by the well known Stern-Gerlach experiment \cite{gerlach1922experimentelle} in  quantum mechanics, and it requires at least 4 hidden states as a lower bound to model this process. \citet{srinivasan2018} demonstrated that HQMMs \emph{learned} from such synthetic data showed in practice the same benefits that held in theory. Our goal is to verify that the COSM method performs at least as well as the GS method on a dataset well-suited to the HQMM model class.

\begin{figure}[h]
 \centering
    \includegraphics[scale = 0.37]{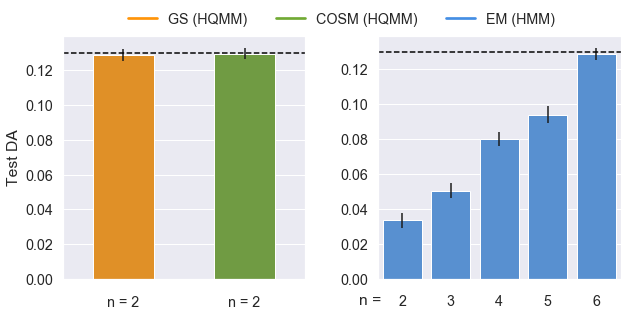}
      \caption{\textbf{Test Set Performance on the Synthetic HQMM Data}: The dashed line represents the test set performance of the true model that generated the data. The GS and COSM algorithms were used to learn ($2,6,1$)-HQMMs, while EM was used to learn HMM models with varying number of hidden states ($n$). A $6-$state HMM model was needed to match a $2-$state HQMM.}
      \label{fig:synthetic_hqmm_results} 
\end{figure}

We used the same synthetic dataset used by \citet{srinivasan2018}, with 20 training and 10 validation sequences of length 3000. We further split up each sequence into 300 sequences and use a burn-in of 100, instead of training on 3000-length sequences with a burn-in of 1000. Splitting up the data in this fashion did not change the amount of training data used, and reduced training time without impacting accuracy. We trained HQMMs using the COSM approach for 60 epochs, and saved the model that yielded the highest DA score on the validation set; we used this model to evaluate on the test set of 10 sequences of length 3000 (with burn-in 1000). For our hyperparameters, we set learning rate $\tau=0.75$ and decay rate $\alpha=0.92$. The results for this model are shown in  Figure~\ref{fig:synthetic_hqmm_results}.

\vspace{-2mm}
We see that the COSM method achieves slightly better DA compared to the GS method. We confirm that as seen in \citet{srinivasan2018}, we need a $6-$state HMM to model this $2-$state HQMM.
\subsection{Synthetic HMM Data}
For our second experiment, we generated data using the same synthetic HMM as \citet{srinivasan2018}, with 6 hidden states and 6 possible outputs. We show two things with the experiments on this dataset: 1) the COSM method is still able to find better optima than the GS method on a dataset \emph{not} generated by a quantum model, and 2) the COSM method is much faster than the previous GS method -- so much so that we are able to train larger HQMMs that previously took too long to train. Finally, we investigate the effects on increasing model size by adding latent states ($n$) versus increasing the dimension of the ancilla variable ($w$).

\begin{figure}[h!]
  \centering
    \includegraphics[scale = 0.39]{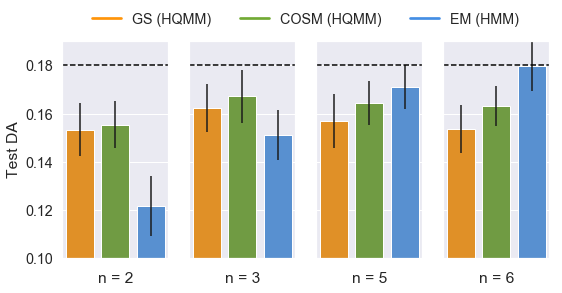}
      \caption{\textbf{Test Set Performances on the Synthetic HMM Data for Models with $w$ = 1}: The dashed line represents the test set DA of the true model (a $(6,6)-$HMM) that generated the data. Test set DAs shown for HQMMs with $w=1$ learned using GS and COSM algorithms and HMMs learned with EM. The number of hidden states ($n$) is varied. The HQMMs learned using COSM consistently outperformed those learned using GS.}
      \label{fig:synthetic_hmm_w1_results}
\end{figure}

\begin{figure}[h!]
  \centering
    \includegraphics[scale = 0.39]{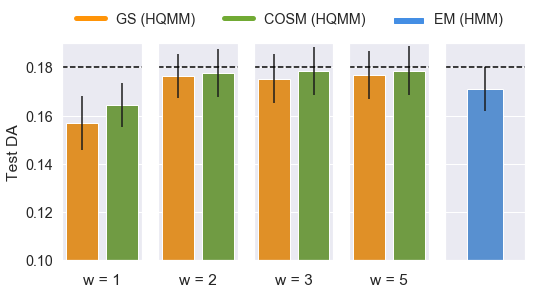}
      \caption{\textbf{Test Set Performances on the Synthetic HMM Data for Models with $n$ = 5}: The dashed line represents the test set performance of the true model (a $(6,6)-$HMM) that generated the data. Test set DAs shown for HQMMs with $n=5$ learned using GS and COSM algorithms and HMMs learned with EM. The dimension of the environment variable ($w$) is varied (does not exist for HMMs). Again, COSM outperforms GS, and increasing $w$ yields performance comparable to the true model and better than the learned HMM with the same number of hidden states.}
      \label{fig:synthetic_hmm_n5_results}
\end{figure}

We used the same 20 training and 10 validation sequences of length 3000 used by \citet{srinivasan2018}, splitting up each sequence into 300 sequences and use a burn-in of 100. We trained HQMMs using the COSM approach for 60 epochs, and evaluated the model with the highest validation DA score on the test set. For our hyperparameters, we set learning rate $\tau=0.75$ and decay rate $\alpha=0.92$. The results for this model are shown in  Figure~\ref{fig:synthetic_hmm_w1_results} and Figure~\ref{fig:synthetic_hmm_n5_results}.

\paragraph{COSM finds better optima than GS} In Figure~\ref{fig:synthetic_hmm_w1_results}, HQMMs (with $w=1$) learned using COSM achieve better optima than HQMMs learned using GS for all $n$. We also confirm that as noted in \citet{srinivasan2018}, small HQMMs ($n < 4$) can model this data better than small HMMs, while for $n > 4$, we see the opposite. Also note that the number of parameters for an HQMM scales faster than for an HMM. In Figure~\ref{fig:synthetic_hmm_n5_results}, we take advantage of the additional hyperparameter $w$ available to HQMMs and present results for $(5,6,w)-$HQMMs (varying $w$). We see that a $(5,6,2)-$HQMM is sufficient to outperform a $(5,6)-$HMM. Furthermore, as we increase $w$, HQMMs approach the DA for the true HMM model (dashed line).

\begin{figure}[h!]
    \centering
    \includegraphics[scale=0.363]{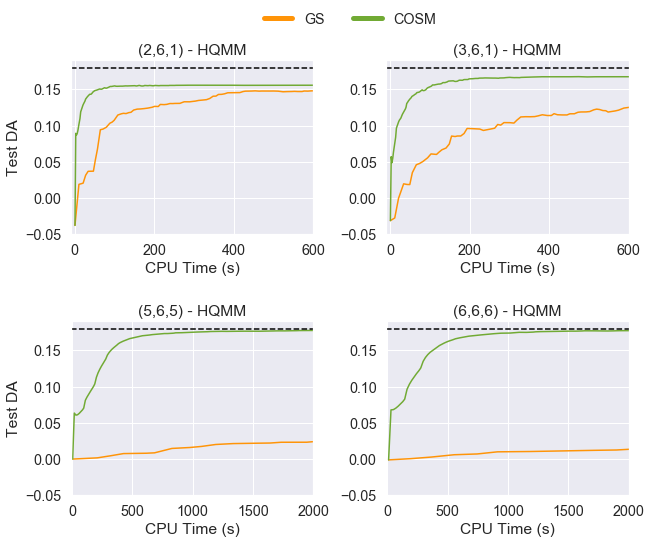}
    \caption{\textbf{COSM Learns More Accurate Models Faster than GS}: Test DA versus training time for various $(n,s,w)$-HQMMs trained on the synthetic HMM data. COSM converges to a better optimum much faster than GS for all models; the dashed line represents the test set DA of the true model that generated the data.}
    \label{fig:hmm_results_training_times}
\end{figure}

\paragraph{COSM is much faster than GS} In Figure \ref{fig:hmm_results_training_times}, we plot the test set DA versus CPU training time for the smallest and largest models trained. To ensure a fair comparison, we train both approaches on sequences of length $300$ and a batch size of 30. Note that we pre-tune hyperparameters on the validation set, and the graphs show the changing test DA as the models are trained with these hyperparameters (test DAs were not used to tune hyperparameters).

\vspace{-2mm}
For all models, we see that COSM converges much faster than GS, and the difference in both speed and accuracy is especially pronounced for the larger models; COSM converges within a few hundred seconds, while GS yields very poor solutions even after 2000 seconds. The substantial improvements in speed offered by COSM allow us to easily train larger models that were were too time-consuming to train using GS. \citet{srinivasan2018} proved that a $(6,6,6)$-HQMM should be sufficient to fully model a $(6,6)$-HMM, but the GS method was too slow to train this model. With COSM, we are able to show that this theoretical guarantee holds in practice. In fact, we find that in practice a $(5,6,5)-$HQMM is sufficient to model our $(6,6)-$HMM!

\begin{figure}[h!]
    \centering
    \includegraphics[scale = 0.42]{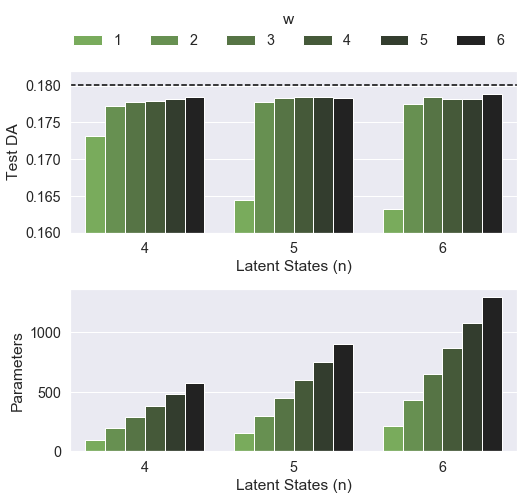}
    \caption{\textbf{Increasing $w$ Instead of $n$ Yields Comparable Performance with Fewer Parameters.} Test set DAs (above) and the number of model parameters (below) for various HQMMs trained on the synthetic $(6,6)$-HMM data using the COSM algorithm. In general, past a certain  value of $n$, we get a larger marginal increase in DA by increasing $w$ as opposed to increasing $n$. The number of model parameters grow linearly with $w$ and quadratically with $n$.
    }
    \label{fig:nwvary}
\end{figure}

\paragraph{Investigating increasing $n$ vs $w$} Finally, we wanted to empirically investigate the properties of scaling the model-size of an HQMM; for a fixed number of outputs, there are two ways to increase the number of parameters: we can either increase the dimension of the latent state $n$, or increase the dimension of the `environment' variable $w$. We present the results for various $n,w$ in Figure \ref{fig:nwvary} (grouped by $n$, although we can compare bars of the same colour across groups to identify change in DA as we change $n$). 

We see that there is a large gain in performance in going from $w=1$ to $w=2$, but the benefits of increasing $w$ past that are less clear. We also see that for fixed $w >  1$, the value of $n$ does not matter much. Note that we count every complex-valued parameter as a single parameter.  The number of parameters of an HQMM scales quadratically with $n$  and linearly with $w$, and the necessary matrix inverse during training only depends on $n$, we should also expect computational benefits for scaling with $w$ over $n$. Further theoretical investigation will be needed to understand the tradeoffs of increasing $n$ versus  $w$.

\subsection{Splice Dataset}
For our final experiment, we use the real-world splice dataset\footnote{https://archive.ics.uci.edu/ml/datasets/Molecular+Biology+(Splice-junction+Gene+Sequences)} consisting of DNA sequences of length 60, each element of which represents a nucleobase \cite{Dua2017}. These DNA nucleobases fall into one of four categories: Adenine (A), Cytosine (C), Guanine (G), and Thyamine (T). A DNA sequence typically consists of information encoded in sub-sequences (exons), that are separated from one another through superfluous sub-sequences (introns). The task associated with this dataset is to classify sequences as having an exon-intron (EI) splice, an intron-exon (IE) splice, or neither (N), with 762, 765, and 1648 labeled examples for each label respectively. In addition to A, C, T and G, the raw dataset also contains some ambiguous characters, which we filter out prior to training. Our goal in this experiment is to demonstrate that we can now train HQMMs on a real-world dataset using COSM; this would have been too slow to train using GS. 

\begin{figure}[h!]
    \centering
    \includegraphics[scale = 0.37]{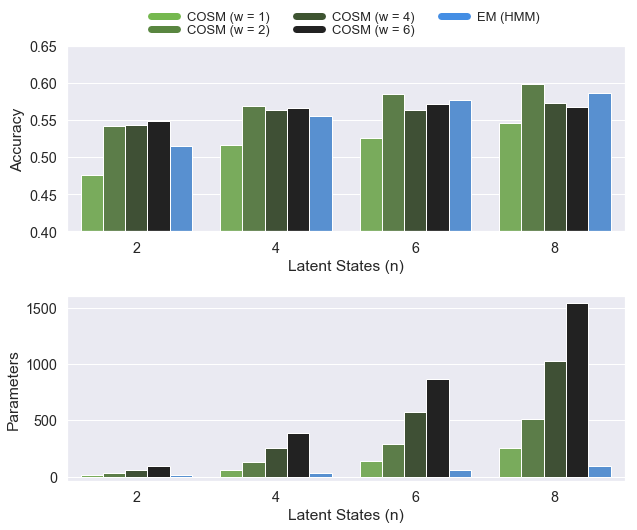}
    \caption{\textbf{Average 5-fold Test Set Performance on the Splice Dataset} Test set accuracies (above) and number of parameters (below) for various HQMM and HMM models trained using the COSM and EM algorithms respectively. Increasing $w$ generally yields greater accuracy than HMM models with the same $n$, albeit with many more parameters.}
    \label{fig:splice_results}
\end{figure}

We train a separate model for each of the three labels, and during test-time, choose the label corresponding to the model that assigned the highest likelihood to the given sequence. We train HQMMs using the COSM method and HMMs with the EM algorithm for reference. In Figure~\ref{fig:splice_results}, we report the average classification accuracies across all labels obtained with 5-fold cross validation. For reference, a random classifier achieves around $33.3\%$ accuracy.

Note that 5-fold cross-validation would have been prohibitively time consuming for the GS algorithm, even for models with a modest number of parameters. However, we are able to learn these  HQMMs with COSM! We also see that, (as before) there is a sizable marginal gain in DA when going from $w=1$ to $w=2$, with the benefits of increasing $w$ further being less clear. However unlike the previous experiment, we still see persistent gains by increasing $n$. Interpreting this in conjunction with the results in the previous section suggests that we have to tune both $n$ and $w$ depending on the dataset.

We also find that for a given number of hidden states, COSM is able to learn an HQMM that outperforms the corresponding HMM, although this comes at the cost of a rapid scaling in the number of parameters. Again, we count every complex-valued parameter as a single parameter. 

\section{Conclusion}
We discussed QGMs, a retraction-based learning algorithm that directly constrains gradient updates to the Stiefel manifold to learn feasible QGMs, and presented experimental results on two synthetic datasets and one real-world dataset. In the process, we showed that the proposed algorithm outperforms the prior approach in terms of both speed and accuracy, and so were able to train HQMMs that were previously too large to train. This also  suggests that directly optimizing the parameters is a better strategy  than finding small, local unitary  rotations of the matrix on the Stiefel manifold. We also found that HQMMs can perform well not only on data generated by quantum models, but on other data as well.
One downside is the rapid scaling of parameters in HQMMs, and it would be interesting to investigate approximations that may produce similar performance with far fewer parameters. Other future work could also investigate the learning problem for more complicated QGMs and provide a more rigorous theoretical basis for understanding how model capacity and expressivity change when we scale HQMMs by $n$ versus $w$. The availability of a fast and scalable learning algorithm for QGMs should open the door for more sustained research in this area. 

\bibliographystyle{apalike}

\begin{thebibliography}{}

\bibitem[Absil et~al., 2007]{Absil2007}
Absil, P.-A., Mahony, R., and Sepulchre, R. (2007).
\newblock {\em Optimization Algorithms on Matrix Manifolds}.
\newblock Princeton University Press, Princeton, NJ, USA.

\bibitem[A.Hj{\o}rungnes and D.Gesbert, 2007]{Hjorungnes2007}
A.Hj{\o}rungnes and D.Gesbert (2007).
\newblock {Complex-Valued Matrix Differentiation: Techniques and Key Results}.
\newblock 55(6):2740--2746.

\bibitem[Clark et~al., 2015]{clark2015hidden}
Clark, L.~A., Huang, W., Barlow, T.~M., and Beige, A. (2015).
\newblock Hidden {Q}uantum {M}arkov {M}odels and {O}pen {Q}uantum {S}ystems
  with {I}nstantaneous {F}eedback.
\newblock In {\em ISCS 2014: Interdisciplinary Symposium on Complex Systems},
  pages 143--151. Springer.

\bibitem[Dheeru and Karra~Taniskidou, 2017]{Dua2017}
Dheeru, D. and Karra~Taniskidou, E. (2017).
\newblock {UCI} machine learning repository.

\bibitem[Edelman et~al., 1998]{Edelman1998}
Edelman, A., Arias, T.~A., and Smith, S.~T. (1998).
\newblock {The Geometry of Algorithms with Orthogonality Constraints}.
\newblock 20(2):303--353.

\bibitem[Gerlach and Stern, 1922]{gerlach1922experimentelle}
Gerlach, W. and Stern, O. (1922).
\newblock Der experimentelle nachweis der richtungsquantelung im magnetfeld.
\newblock {\em Zeitschrift f{\"u}r Physik}, 9(1):349--352.

\bibitem[Gheondea, 2010]{Gheondea2010}
Gheondea, A. (2010).
\newblock {The three equivalent forms of completely positive maps on matrices}.
\newblock {\em Annals of the University of Bucharest}.

\bibitem[Jaeger, 2000]{jaeger2000observable}
Jaeger, H. (2000).
\newblock Observable operator models for discrete stochastic time series.
\newblock {\em Neural computation}, 12(6):1371--1398.

\bibitem[Kraus, 1971]{Kraus1971}
Kraus, K. (1971).
\newblock General state changes in quantum theory.
\newblock {\em Annals of Physics}, 64(2):311--335.

\bibitem[Leifer and Poulin, 2008]{leifer2008quantum}
Leifer, M.~S. and Poulin, D. (2008).
\newblock Quantum graphical models and belief propagation.
\newblock {\em Annals of Physics}, 323(8):1899--1946.

\bibitem[Leifer and Spekkens, 2013]{leifer2013towards}
Leifer, M.~S. and Spekkens, R.~W. (2013).
\newblock Towards a formulation of quantum theory as a causally neutral theory
  of {B}ayesian inference.
\newblock {\em Physical Review A}, 88(5):052130.

\bibitem[M.~Zhao, 2007]{noomreport}
M.~Zhao, H.~J. (2007).
\newblock Norm observable operator models.
\newblock Technical report, Jacobs University.

\bibitem[Monras et~al., 2010]{monras2010hidden}
Monras, A., Beige, A., and Wiesner, K. (2010).
\newblock Hidden {Q}uantum {M}arkov {M}odels and non-adaptive read-out of
  many-body states.
\newblock {\em arXiv preprint arXiv:1002.2337}.

\bibitem[Nielsen and Chuang, 2010]{nielsen_chuang_2010}
Nielsen, M.~A. and Chuang, I.~L. (2010).
\newblock {\em Quantum Computation and Quantum Information: 10th Anniversary
  Edition}.
\newblock Cambridge University Press.

\bibitem[Oza et~al., 2009]{Oza2009}
Oza, A., Pechen, A., Dominy, J., Beltrani, V., Moore, K., and Rabitz, H.
  (2009).
\newblock {Optimization search effort over the control landscapes for open
  quantum systems with Kraus-map evolution}.
\newblock {\em Journal of Physics A: Mathematical and Theoretical}, 42(20).

\bibitem[Pillis, 1967]{Pillis1967}
Pillis, J. (1967).
\newblock {{L}inear {T}ransformations which {P}reserve {H}ermitian and
  {P}ositive {S}emidefinite {O}perators}.
\newblock {\em Pacific Journal of Mathematics}.

\bibitem[Qian, 1999]{Qian1999}
Qian, N. (1999).
\newblock On the momentum term in gradient descent learning algorithms.
\newblock {\em Neural Netw.}, 12(1):145--151.

\bibitem[Rumelhart et~al., 1988]{Rumelhart1998}
Rumelhart, D.~E., Hinton, G.~E., and Williams, R.~J. (1988).
\newblock Neurocomputing: Foundations of research.
\newblock chapter Learning Representations by Back-propagating Errors, pages
  696--699. MIT Press, Cambridge, MA, USA.

\bibitem[Srinivasan et~al., 2018a]{NIPS2018_8235}
Srinivasan, S., Downey, C., and Boots, B. (2018a).
\newblock Learning and {I}nference in {H}ilbert space with {Q}uantum
  {G}raphical {M}odels.
\newblock In {\em Advances in Neural Information Processing Systems 31}.

\bibitem[Srinivasan et~al., 2018b]{srinivasan2018}
Srinivasan, S., Gordon, G., and Boots, B. (2018b).
\newblock Learning {H}idden {Q}uantum {M}arkov {M}odels.
\newblock In {\em International Conference on Artificial Intelligence and
  Statistics}, pages 1979--1987.

\bibitem[Stiefel, 1936]{stiefel-original}
Stiefel, E. (1935-1936).
\newblock Richtungsfelder und fernparallelismus in n-dimensionalem mannig
  faltigkeiten.
\newblock {\em Commentarii Math. Helvetici}, 8(305-353).

\bibitem[Verstraete et~al., 2004]{Verstraete2004}
Verstraete, F., Garc\'ia-Ripoll, J.~J., and Cirac, J.~I. (2004).
\newblock Matrix product density operators: Simulation of finite--temperature
  and dissipative systems.
\newblock {\em Phys. Rev. Lett.}, 93(20):207204.

\bibitem[Warmuth and Kuzmin, 2014]{warmuth2014bayesian}
Warmuth, M.~K. and Kuzmin, D. (2014).
\newblock A {B}ayesian probability calculus for density matrices.
\newblock {\em arXiv preprint arXiv:1408.3100}.

\bibitem[Wen and Yin, 2013]{Wen2013}
Wen, Z. and Yin, W. (2013).
\newblock {A feasible method for optimization with orthogonality constraints}.
\newblock {\em Mathematical Programming}, 142(1-2):397--434.

\bibitem[Yeang, 2010]{yeang2010probabilistic}
Yeang, C.-H. (2010).
\newblock A probabilistic graphical model of quantum systems.
\newblock In {\em Machine Learning and Applications (ICMLA), 2010 Ninth
  International Conference on}, pages 155--162. IEEE.

\end{thebibliography}

\end{document}